\documentclass{article}

\usepackage[nonatbib,preprint]{neurips_2020}

\usepackage[sort&compress,numbers]{natbib}
\usepackage[utf8]{inputenc} 
\usepackage[T1]{fontenc}    
\usepackage{hyperref}       
\usepackage{url}            
\usepackage{booktabs}       
\usepackage{amsfonts}       
\usepackage{nicefrac}       
\usepackage{microtype}      
\usepackage{graphicx}
\usepackage{amsmath,amsfonts,amssymb}
\usepackage{bbm}
\newcommand{\diag}{\operatorname{diag}}
\DeclareMathOperator{\Tr}{Tr}
\title{Thalamocortical motor circuit insights for more robust hierarchical control of complex sequences}

\author{
Laureline Logiaco\thanks{Zuckerman Institute and Department of Psychiatry, Columbia University} \\ \texttt{ll3041@columbia.edu}
\And G.~Sean Escola\footnotemark[1] \\ \texttt{gse3@columbia.edu} \\
}

\begin{document}

\maketitle

\begin{abstract}

We study learning of recurrent neural networks that produce temporal sequences consisting of the concatenation of re-usable ‘motifs’. In the context of neuroscience or robotics, these motifs would be the motor primitives from which complex behavior is generated. Given a known set of motifs, can a new motif be learned without affecting the performance of the known set and then used in new sequences without first explicitly learning every possible transition? Two requirements enable this: \emph{(i)} parameter updates while learning a new motif do not interfere with the parameters used for the previously acquired ones; and \emph{(ii)} a new motif can be robustly generated when starting from the network state reached at the end of any of the other motifs, even if that state was not present during training. We meet the first requirement by investigating artificial neural networks (ANNs) with specific architectures, and attempt to meet the second by training them to generate motifs from random initial states. We find that learning of single motifs succeeds but that sequence generation is not robust: transition failures are observed. We then compare these results with a model whose architecture and analytically-tractable dynamics are inspired by the motor thalamocortical circuit, and that includes a specific module used to implement motif transitions. The synaptic weights of this model can be adjusted without requiring stochastic gradient descent (SGD) on the simulated network outputs, and we have asymptotic guarantees that transitions will not fail. Indeed, in simulations, we achieve single-motif accuracy on par with the previously studied ANNs and have improved sequencing robustness with no transition failures. Finally, we show that insights obtained by studying the transition subnetwork of this model can also improve the robustness of transitioning in the traditional ANNs previously studied.

\end{abstract}

\section{Introduction}

How to engineer systems to perform complex and structured behavior in physical environments has been a long-standing goal of robotics and of artificial intelligence for many decades~\cite{Brooks1986,Prescott1999,Merel2019}. This interest was rewarded with early progress in the design of optimal feedback controllers for a well-defined action in a simple dynamical system, but significant limitations were encountered for controlling more complex systems over a wide range of actions~\cite{Todorov2002}. With the advances in recurrent neural network (RNN) training, this question has been revisited -- notably within the field of deep reinforcement learning~\cite{Heess2017} -- and increased computational power now permits the control of more complex systems. However, significant obstacles remain. First, it remains unclear how to design hierarchically organized modules to interact efficiently to control structured behavior. 
Second, in the setting of acquiring knowledge in a sequential fashion, ANNs suffer from the phenomenon of catastrophic forgetting, where the incorporation of new information may result in the loss of what has been previously learned. Third, ANNs frequently struggle to apply their knowledge flexibly and thus 
rapidly improvise to novel situations~\cite{Merel2019,Sodhani2020}.
Here, we investigate these questions by designing modular network architectures that can avoid catastrophic forgetting, and, in the setting of a simple sequence generation task, we explore network training techniques that permit networks to generalize. Interestingly, we find that an analytically tractable model inspired by the motor thalamocortical architecture provides a robust solution in our task as well as offers insights that we show can improve the performance of gradient-trained networks.

\section{Task and architecture design\label{seq:TasksModels}}

\begin{figure}[t!]

\centering

\includegraphics[width=5in]{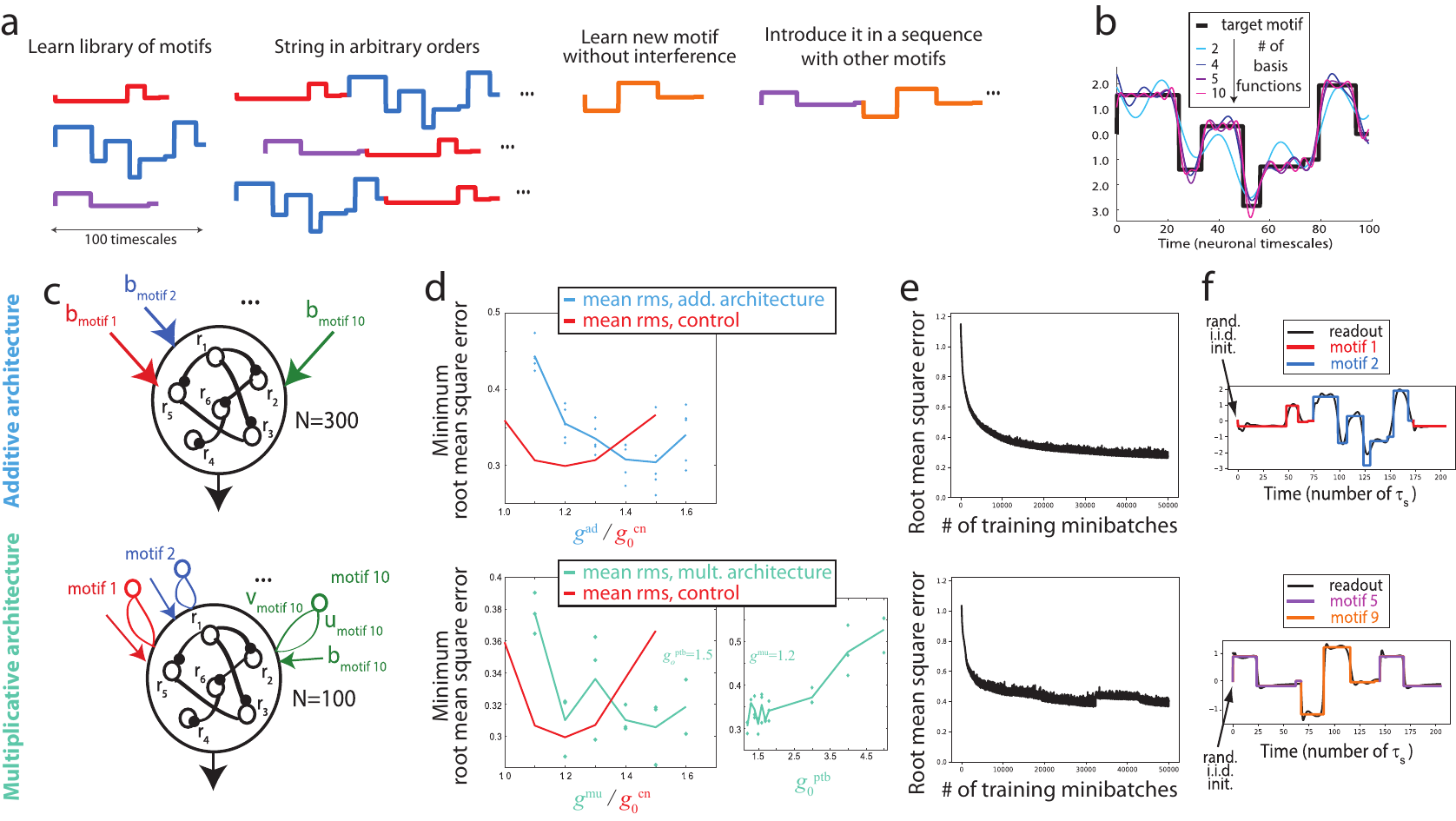}

\caption{\label{fig:TaskArchit} \textbf{Task and candidate networks.} \textbf{a}) Flexible and extendable motif sequencing. Our goals are to execute time-varying motifs from a library in sequences of arbitrary order without training all possible transitions, and to learn motifs without interfering with the existing library for incorporation into novel sequences.
\textbf{b}) Example motif fit with an increasing number of complex exponentials (the eigenmodes of linear dynamical systems). 
\textbf{c-f}) Top row: Additive architecture; bottom row: multiplicative architecture. \textbf{c}) RNNs that may succeed in the task shown in \textbf{a} because they segregate parameters into motif-specific sets (schematized in colors) while benefiting from fixed shared parameters (schematized in black). 
\textbf{d-f}) Hyperparameter optimization when all transitions are trained.
\textbf{d}) Minimum root mean square error over training, depending on the gain hyperparameters $g^{\textrm{ad}}$, $g^{\textrm{mu}}$, and $g_0^{\textrm{ptb}}$ (defined in section~\ref{seq:TasksModels}). Dots are individual networks, the line is the average. In red, for reference, we show the mean minimum root mean square error in the control architecture (averaged over five individually trained networks for each $g_0^{\textrm{cn}}$).
\textbf{e}) Example learning curves.
\textbf{f}) Example training trials (2 of 90 motif pairs that compose a minibatch, sampling all transitions).
}
\end{figure}

Our goal is to study the ability of RNNs to capture two important attributes of successful hierarchical motor control: (\emph i) the ability to acquire a library of complex motor motifs and then learn new motifs without interfering with the execution of previously acquired ones, and (\emph{ii}) the ability to flexibly string motifs into a sequence of arbitrary order without having necessarily rehearsed all of the transitions composing the sequence (Fig.~\ref{fig:TaskArchit}a).
Thus, we will train RNNs with different architectures to learn a set of motifs without interference and test them to generate both learned and novel sequences.

Our target motifs are chosen to challenge the expressivity of continuous-dynamics networks. Specifically, they have discrete jumps at random times but are constant between these jumps (Fig.~\ref{fig:TaskArchit}a and see appendix for the full list of motifs used in this paper). To output both jumps and constant periods, networks must generate high and low frequency oscillations, as is clearly illustrated by the Fourier series decomposition of a square wave or by examining the fit of one of our motifs with different numbers of exponentially modulated sines (such functions being the natural basis functions of linear recurrent networks; Fig.~\ref{fig:TaskArchit}b). 
Furthermore, training RNNs to generate motifs such as these is difficult because of instability issues when gradients are propagated through many recurrent steps~\cite{Pascanu2012,Bengio1994}.

To minimize interference between motifs, we only consider
network architectures
that separate the trainable parameters into motif-specific sets (Fig.~\ref{fig:TaskArchit}c). 
Each architecture employs a shared linear readout and recurrent network (table~\ref{table_params}), but differs in the use of the motif-specific parameters.

In the `additive' architecture (Fig.~\ref{fig:TaskArchit}c, top),
each motif $\mu$ is produced in response to a learned input vector $\mathbf{b}_{\mu}$, leading to the following dynamics for the activities $\mathbf{x}$:

\begin{equation}
\tau \; \dot{\mathbf{x}} = - \mathbf{x} + g^{\textrm{ad}} \, \mathbf{J} \tanh \left(
\mathbf{x} \right) + \mathbf{b}_{\mu},
\end{equation}

where the gain $g^{\textrm{ad}}$ is a hyperparameter and $\mathbf{J}$ is the connectivity matrix -- with iid elements taken from a centered Gaussian with standard deviation (std) $1/\sqrt{N}$ (previous work has shown that if $g^{\textrm{ad}}>1$ this leads to a rich dynamical regime appropriate for complex computations~\cite{Sompolinsky1988,Sussillo2009}).

In the `multiplicative' architecture (Fig.~\ref{fig:TaskArchit}c, bottom; inspired by both previous machine learning literature~\cite{ICML2011Sutskever} and the motor thalamocortical architecture~\cite{Logiaco2019}), each motif $\mu$ is produced in response to both a learned input vector $\mathbf{b}_{\mu}$ and a learned rank-one perturbation of the connectivity $\mathbf{u}_{\mu} \mathbf{v}^{\intercal}_{\mu}$. The latter is equivalent to a loop through an instantaneous `unit' receiving input from the recurrent network through the weights $\mathbf{v}_{\mu}$ and feeding back through the weights $\mathbf{u}_{\mu}$. The dynamics are then:

\begin{equation}
\tau \; \dot{\mathbf{x}} = - \mathbf{x} + \, \left(  g^{\textrm{mu}} \, \mathbf{J} + \mathbf{u}_{\mu} \mathbf{v}^{\intercal}_{\mu} \right) \tanh \left(
\mathbf{x} \right) + \mathbf{b}_{\mu},
\end{equation}

where $g^{\textrm{mu}}$ and $\mathbf{J}$ are defined as for the additive network, and $\mathbf{u}_{\mu}$ and $\mathbf{v}_{\mu}$ are each learned and initialized iid from a centered Gaussian with std $g_0^{\textrm{ptb}}/\sqrt{N}$ (i.e. expected norm $g_0^{\textrm{ptb}}$).

Finally, we also consider a `control' network that does not separate all trainable parameters into motif-specific sets. Our aim was to assess to what extent lifting this constraint can result in performance gains. This control network was inspired by work showing that -- if hyperparameters are properly optimized -- performance appears to be relatively independent of the architecture and only dependent on the number of parameters that are directly trained~\cite{Collins2017}. Therefore, for our control network, we trained all input, recurrent, and output weights with the number of neurons adjusted such that this network has as many tunable parameters as the additive and multiplicative architectures for a fixed number of motifs (10 motifs; see table~\ref{table_params}). The dynamics of this control network are:
$
\tau \; \dot{\mathbf{x}} = - \mathbf{x} +  \mathbf{J} \tanh \left(
\mathbf{x} \right) + \mathbf{b}_{\mu},
$
where $\mathbf{J}$ is initialized with iid elements taken from a centered Gaussian with std $g_0^{\textrm{cn}}/\sqrt{N}$.

In all networks, the output $y$ is produced through a linear combination of the rates $\mathbf{r}=\tanh\mathbf{x}$: $y=\mathbf w^\intercal\mathbf r$. The elements of $\mathbf w$ are sampled iid from a centered Gaussian distribution with std $1/\sqrt{N}$ and are either \emph{(i)} fixed (additive and multiplicative networks) or \emph{(ii)} optimized during training (control network). Given the $\tanh$ nonlinearity, such a readout vector can produce network outputs of maximal positive or negative magnitude equal to $\sqrt{N}$, which is sufficient for generating all of our motifs.

\begin{table}
  \caption{Number of neurons and number of parameters in the different architectures}
  \label{table_params}
  \centering
  \begin{tabular}{llll}
    \toprule
    & Additive & Multiplicative & Control \\
    \midrule
  \# of recurrent units $N$ & 300 & 100 & 50 \\
    \midrule
  \# of learned parameters for 10 motifs & 3000 & 3000 & 3050 \\
    \midrule
  \# of motif-specific parameters per motif & 300 & 300 (input: 100, loop: 200) & 50  \\   
    \bottomrule
  \end{tabular}
\end{table}

\section{Gradient-trained RNNs trade off flexibility and robustness}\label{sec:gradientRNNresults}


\subsection{Optimizing learning using a training set with all possible transitions}

We trained these RNNs using gradient descent (more specifically, ADAM~\cite{Kingma2015}) with a time discretization of $dt=0.1 \tau$ and $\tau=1$ such that each motif is roughly a thousand timesteps (appendix). Our objective function was the mean square error between desired and actual output.
After testing various parameters of ADAM, we identified the following as yielding successful training in our setting: $\textrm{learning rate}=10^{-4}$, $\beta_1=\beta_2=0.5$, and $\epsilon=10^{-8}$.
In order to verify the basic ability of our networks to perform motif sequencing, as well as to have a reliable performance criterion with which to choose appropriate hyperparameter values (i.e., $g^{\textrm{ad}}$, $g^{\textrm{mu}}$, $g_0^{\textrm{ptb}}$, and $g_0^{\textrm{cn}}$), we first trained our networks on all possible transitions between motifs: 90 trials consisting of a sequence of at least two motifs (optionally followed by the start of the first motif such that all trials were of equal duration). The sequences were initialized with network activities sampled from the standard normal distribution. Example trials are shown in Fig.~\ref{fig:TaskArchit}f. We trained the networks over 50,000 minibatches, after which the derivative of the loss function was near zero (Fig.~\ref{fig:TaskArchit}e).
We repeated training many times to optimize  $g^{\textrm{ad}}$, $g^{\textrm{mu}}$, $g_0^{\textrm{ptb}}$, and $g_0^{\textrm{cn}}$. Similarly to previous reports~\cite{Collins2017}, we find that after sweeping through these hyperparameters, the peak performance for each architecture appears to be approximately equivalent (Fig.~\ref{fig:TaskArchit}d). Interestingly, the multiplicative architecture -- inspired by the thalamocortical model~\cite{Logiaco2019} that we will evaluate in the next section -- appears to be less sensitive to choices of the gain parameters. For this multiplicative architecture, theoretical arguments suggest that the loop weights $\mathbf u_\mu$ and $\mathbf v_\mu$ might need to be large to have a major impact on the recurrent network dynamics~\cite{Schuessler2020,Tao2011,Logiaco2019}. Therefore, we tried initializing $g_0^{\textrm{ptb}}$ to large values, but this appeared to hurt rather than help performance (Fig.~\ref{fig:TaskArchit}d, bottom right). Of note, while the learning curves for our additive RNNs usually appeared relatively smooth, the learning curves for our multiplicative RNNs often included multiple quasi-plateaus followed by more rapid cost function decreases, suggesting the presence of slow points or local minima in the stochastic cost function (Fig.~\ref{fig:TaskArchit}e).

\subsection{Training without transitions yields transition failures upon testing}

\begin{figure}[t!]

\centering

\includegraphics[width=5in]{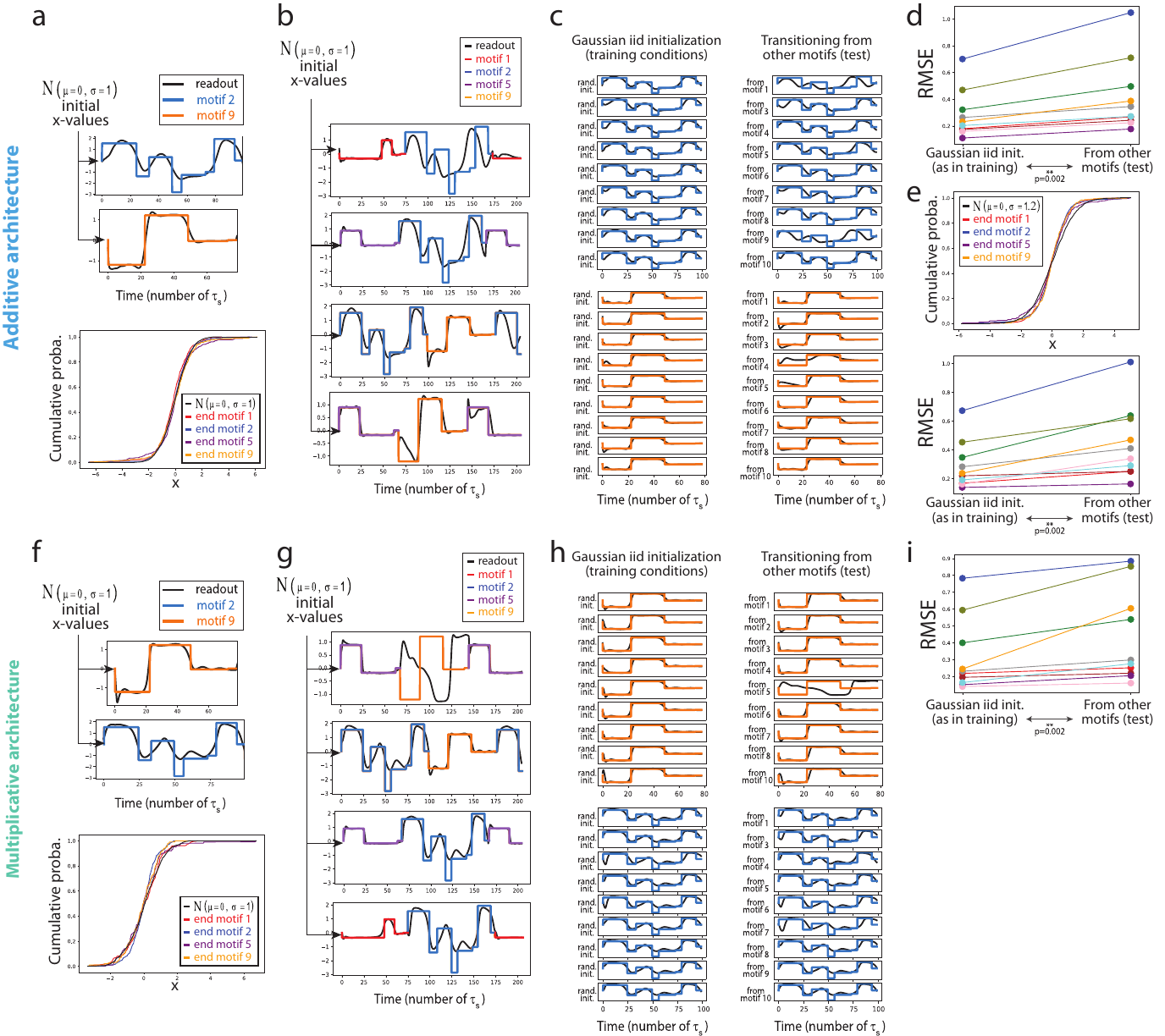}

\caption{\label{fig:ANN_lim_robutsness} \textbf{Transition failures in gradient-trained RNNs}. \textbf{a-e}) Additive architecture; \textbf{f-i}) multiplicative architecture. \textbf{a,f}) Top: Example training trials with single motifs starting from iid standard Gaussian initial $\mathbf x$ values. Bottom: marginal distribution of $\mathbf x$ values, comparing the initial conditions during training with examples of end-of-motif distributions. \textbf{b,g}) Example test trials. \textbf{c,h}) Comparing motif production when starting with iid standard Gaussian initial $\mathbf x$ values (left) as opposed to when preceded by other motifs during sequences (right). \textbf{d,i}) Change of root mean square error, separately for the 10 motifs (color coded as in the other panels), between the training condition (starting from standard Gaussian iid initial $\mathbf x$ values) and the sequencing condition. For each motif, this corresponds to the root mean square error on the left \emph{vs} right column of panels \textbf{c} and \textbf{h}. The p-values are for a two-sided Wilcoxon signed-rank test. \textbf{e}) For the additive architecture, training single motifs starting from Gaussian iid $\mathbf x$ values with zero mean but standard deviation of 1.2. Conventions as in panels \textbf{a} and \textbf{d}.}
\end{figure}

The results in the prior section employed a training set with all pairs of transitions represented. This approach requires that, when adding a new motif to a network's behavioral library, the transitions to and from all previously acquired motifs have to be learned. This ultimately scales quadratically with the number of motifs and is thus prohibitive as a solution for extensible and flexible motor sequencing. As a possible solution, here we propose training each motif in isolation but with initial network activities selected to approximate the distribution of activities at the ends of all other motifs. This solution attempts to leverage the ability of RNNs to be trained to process samples from a known distribution and to then be able to generalize when presented with samples from a similar distribution (e.g.~\cite{VezhnevetsFeudal2017}). In our scenario, if we can approximate the distribution of end-of-motif activities well enough and if training succeeds, then all transitions will work with no transition-specific training. Unfortunately, this distribution is unknown and is training-dependent. However, since we are using networks with large $N$, Gaussian weights, and a $\tanh$ nonlinearity, a Gaussian should well-approximate the marginal statistics of this unknown distribution~\cite{Landau2018,Rivkind2017}. Indeed, we observe that a standard normal is a good choice (Fig.~\ref{fig:ANN_lim_robutsness}a,f,e).

Hence, here our training set consists of single motifs starting with standard normal iid entries for $\mathbf{x}$. We set our hyperparameters to the optima observed in Fig.~\ref{fig:TaskArchit}d ($g^{\textrm{ad}}=1.4$, $g^{\textrm{mu}}=1.4$ and $g_0^{\textrm{ptb}}=1.5$) and otherwise trained as above. For both network types, training was successful (Fig.~\ref{fig:ANN_lim_robutsness}a and f).

We then asked our networks to generate sequences (examples are shown in Fig.~\ref{fig:ANN_lim_robutsness}b and g) and found that even though many transitions are successful, there are cases where the performance of a given motif is substantially degraded throughout its entire duration when preceded by a particular other motif. This is a transition failure.
Interestingly, the shape of the marginal distribution of the elements of $\mathbf x$ at the end of a motif was not a good predictor of whether transitions from this motif would lead to worse performance. For instance, in the case of the additive network, the transition from motif 1 to motif 2 leads to poor performance on motif 2 (Fig.~\ref{fig:ANN_lim_robutsness}b, top), even though the distribution of $\mathbf x$ at the end of motif 1 appears extremely similar to a standard Gaussian (Fig.~\ref{fig:ANN_lim_robutsness}a, bottom). Conversely the distribution of $\mathbf x$ at the end of motif 5 is wider than the standard Gaussian but it does not lead to a bad transition to motif 2 (Fig.~\ref{fig:ANN_lim_robutsness}b, second row). In Fig.~\ref{fig:ANN_lim_robutsness}c and h, we further illustrate performance for two example motifs when transitioning from all other other motifs as opposed to when starting from multiple samples of a standard Gaussian as during training. The statistics for the root mean square errors over these random \emph{vs.} end-of-other-motif initial conditions are shown in Fig.~\ref{fig:ANN_lim_robutsness}d,i. For all the 10 motifs that were learned, the root mean square error increased when transitioning, a significant result according to a two-sided Wilcoxon signed-rank test (p=0.002). 

\begin{figure}[t!]

\centering

\includegraphics[width=5in]{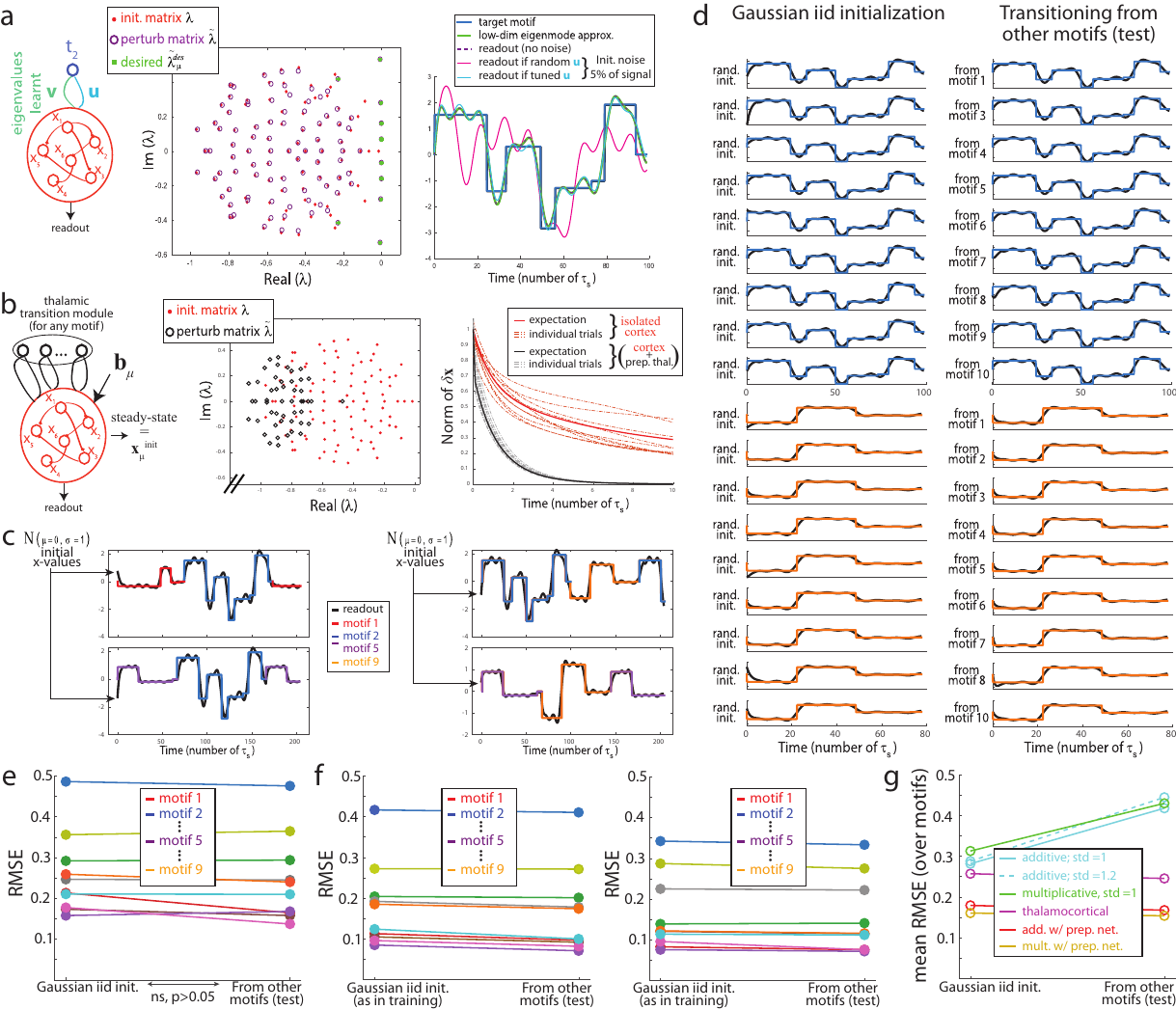}

\caption{\label{fig:thalamocort} \textbf{Robust transitions in thalamocortical model}. \textbf{a}) Adjusting a motif-specific loop through the thalamic unit $\textbf{t}_2$ (left), leading to the control of both eigenvalues (middle) and eigenvectors of the dynamics $\mathbf{x}(t)$ such that the readout robustly follows motif 2 (right). \textbf{b}) Thalamic module used for all motif transitions (left). After optimization, the thalamic module sets the eigenvalues of the dynamics to be more negative (middle) which contributes to a fast decrease of the distance to steady-state $|\delta\mathbf x|$ (right). \textbf{c}) Example sequences. \textbf{d}) As in Fig.~\ref{fig:ANN_lim_robutsness}c,h, for the thalamocortical model. \textbf{e}) Change of root mean square error for each motif between random initialization and sequencing conditions (not significant as per a Wilcoxon signed rank test). \textbf{f}) As in \textbf e, but for the additive (left) and multiplicative (right) networks when augmented with a thalamic transition module for the first $5\tau$ of each motif. \textbf{g}) For the different network architectures, the average root mean square error in both the random initialization and sequencing conditions.}
\end{figure}

Figs.~\ref{fig:ANN_lim_robutsness}a,f shows that the std of the marginal distribution of the values of $\mathbf x$ slightly exceeds 1 for some motifs. To ensure that the transition failures we observed when initializing with the standard normal were not due to this, we retrained our additive network with initial $\mathbf x$ values sampled from a Gaussian with std of 1.2 (Fig.~\ref{fig:ANN_lim_robutsness}e). We observe that the end-of-motif $\mathbf x$ distributions still have standard deviations around 1 for all motifs (Fig.~\ref{fig:ANN_lim_robutsness}e top). Despite training with a wider distribution, performance errors after transitioning still occurred with motif production even more impaired compared to training with a standard normal (Fig.~\ref{fig:ANN_lim_robutsness}e bottom and Fig.~\ref{fig:thalamocort}f).

The failure to robustly transition between motifs when training on single motifs that are randomly initialized is a particular instance of an out-of-distribution generalization limitation. Not only is the true marginal distribution of each element of $\mathbf x$ at the end of a given motif not exactly matched to the Gaussian distribution we use during training, but additionally these elements can be highly correlated with each other due to correlated inputs and the recurrent structure of the network. This problem is very hard to solve if we do not want to train all possible transition pairs since these correlations are instrumental for performance and we do not know them in advance before training.

As we are unable to implement robust and flexible sequencing with classical techniques used to train RNNs, we now investigate a different potential solution that leverages insights from the motor thalamocortical circuit. These insights constrain not only the network architecture, but also the specific dynamical regimes of the network during different parts of a motif sequence.

\section{Flexible, robust and extensible sequencing in a thalamocortical model}\label{sec:TCmodel}

The thalamocortical model we investigate follows prior work described by Logiaco and colleagues~\cite{Logiaco2019}, but improves on
this to ensure that the transitions between motifs are smooth.  

\subsection{Model construction and parameter adjustment}
The model parameters can be adjusted through analytical and semi-analytical techniques which do not require stochastic gradient descent on the simulated dynamics, a major advantage compared to standard training of ANNs. Details are described elsewhere~\cite{Logiaco2019} and we only provide a brief overview here.
The model consists of a recurrent cortical module with connectivity $\mathbf{J^\mathrm{cc}}$ and activities $\mathbf{x}$ whose projection through the readout weights $\mathbf{w}$ constitutes the output. This cortical module interacts with a non-recurrent thalamic module through instantaneous loops consisting of corticothalamic and thalamocortical weights. We model the basal ganglia, which provides inhibitory input into thalamus, as selectively disinhibiting specific thalamic loops in order to cause execution of the associated motif. Thus, the entire model is a switching linear system.

\textbf{Motif execution:} During motif $\mu$, a single thalamic loop is disinhibited leading to the dynamics:
\begin{align}
    \tau \dot{\mathbf x} & = \tilde{\mathbf J}_\mu \mathbf x,
    \quad\mbox{where}\quad
    \tilde{\mathbf J}_\mu\equiv g \left(\mathbf J^\mathrm{cc} - \mathbf{I} \right) + \mathbf{u}_\mu \mathbf{v}_\mu^\intercal,
    \label{eq:InitGeneRateDyn}
\end{align}

with motif-specific loop vectors $\mathbf u_\mu$ and $\mathbf v_\mu^\intercal$. 
We now consider how these dynamics can approximate a desired output $y_\mu$, knowing that in general a good approximation for $y_\mu$ can be reached through a linear combination of a small number $K$ of complex exponentials: $y_\mu(t)\approx\hat{y}_\mu(t)=\sum^K_{k=1}  [\hat{\boldsymbol\alpha}_\mu]_{k} e^{[\hat{\boldsymbol\lambda}_\mu]_kt}$ (Fig.~\ref{fig:TaskArchit}b,~\cite{Logiaco2019}). The cortical readout can exactly match $\hat{y}_\mu$ if the eigenvalues of $\tilde{\mathbf J}_\mu$ contain the entries of the vector $\hat{\boldsymbol\lambda}_\mu$ and if the initial network activities $\mathbf x^{\mathrm{init}}_\mu$ are set correctly. We accomplish the former (Fig.~\ref{fig:thalamocort}a) by setting $\mathbf v_\mu=\mathbf L^\intercal\diag(\mathbf{Lu_\mu})^{-1}\mathbf Q^+ \mathbf{1},$ where $\mathbf{L}$ is the left eigenvector matrix of $\mathbf{J^\mathrm{cc}}$, and $Q_{k,j}=1/([\hat{\boldsymbol\lambda}_\mu]_k-\lambda_{j})$ where $\lambda_{j}$ is an eigenvalue of $g\left( \mathbf{J^\mathrm{cc}} - \mathbf{I} \right)$. Next, we set the initial activities at the beginning of motif $\mu$ to $\mathbf{x}_\mu^\textrm{init}= \tilde{\mathbf R}_{\mu} \, \diag(\tilde{\mathbf R}_{\mu}^{\intercal} \mathbf{w})^{-1} \boldsymbol{\alpha}_{\mu}$ where $\tilde{\mathbf R}_{\mu}$ contains right eigenvectors of $\tilde{\mathbf J}_\mu$ (with the first $K$ columns corresponding to the eigenvalues in $\hat{\boldsymbol\lambda}_\mu)$, and $[\boldsymbol{\alpha}_{\mu}]_{k\le K} = [\hat{\boldsymbol{\alpha}}_{\mu}]_k$ and $[\boldsymbol{\alpha}_{\mu}]_{k>K} =0$.

The preceding two steps do not specify $\mathbf{u}_\mu$, and with random $\mathbf{u}_\mu$ the readout will be highly sensitive to noise in $\mathbf x^{\mathrm{init}}_\mu$ (pink trace in Fig.~\ref{fig:thalamocort}a right; see \cite{Logiaco2019}). However, if $\mathbf{u}_\mu$ is set to minimize the analytically-computed expected readout deviation due to noise in $\mathbf x^{\mathrm{init}}_\mu$, then robust readout is possible (cyan trace in Fig.~\ref{fig:thalamocort}a right; minimization of the cost $C\left( \mathbf{u} \right)$ in~\cite{Logiaco2019}).

\smallskip
\textbf{Motif transitions:} To successfully transition to motif $\mu$, it is sufficient to implement a mechanism by which $\mathbf x$ approaches $\mathbf x^{\mathrm{init}}_\mu$, which will be the case if the dynamics during a so-called "preparatory period" has $\mathbf x^{\mathrm{init}}_\mu$ as its steady-state. Additionally, it is desirable that the transition dynamics are fast and that they do not cause large transients values on the readout while relaxing to steady-state~\cite{Kaufman2014}. To achieve this, we employ a specific thalamic subpopulation of size $P$ which is disinhibited during all motif transitions, as well as a constant input $\mathbf b_\mu$ specific to the upcoming motif $\mu$ (which here, unlike for the ANNs above, is only present during the transition periods), leading to the dynamics:
\begin{align}
\tau \dot{\mathbf x}= \mathbf J_\textrm{prep}\mathbf x+\mathbf b_\mu,
\quad\mbox{where}\quad
    \mathbf J_\textrm{prep} \equiv g \left( \mathbf{J^\mathrm{cc}} - \mathbf{I} \right) + \mathbf U_{\textrm{prep}} \mathbf V_{\textrm{prep}}^\intercal,
\end{align}
with $N\times P$ preparatory loop weights $\mathbf U_{\textrm{prep}}$ and $\mathbf V_{\textrm{prep}}$. With these dynamics, the activity at steady-state will match $\mathbf{x}^{\textrm{init}}_\mu$ if $\mathbf b_\mu= -\mathbf J_\textrm{prep} \mathbf{x}^{\textrm{init}}_\mu$ (Fig.~\ref{fig:thalamocort}b).
Note that the difference $\delta\mathbf x$ between the cortical activities and their steady state decays at a rate that is independent of $\mathbf{b}_\mu$ and therefore of the upcoming motif: $\tau \dot{\delta\mathbf x} = \mathbf J_\textrm{prep}\delta\mathbf x$ for all $\mu$. This allows us to optimize the same weights $\mathbf U_{\textrm{prep}}$ and $\mathbf V_\textrm{prep}$ to favor rapid and smooth transitions between all pairs of motifs. Following ref.~\cite{Logiaco2019}, we achieve fast transitions by minimizing the time-integral of the expected square norm of $\delta\mathbf x$, with rates $\delta\mathbf x_0$ at the beginning of the transition period sampled iid. We also augment our cost function with the time-integral of the expected squared derivative of the readout to ensure smooth transitions. Our total cost function is therefore:
\begin{align}
    C(\mathbf U_\textrm{prep},\mathbf V_\textrm{prep}) &=  E_{ \delta\mathbf x_0 } \left[ \int_0^\infty\!\! dt \left\| \delta\mathbf x \right\|^{2} \right] + \beta \, N \, E_{ \delta\mathbf x_0 } \left[ \int_0^\infty\!\! dt  \left(\frac{d}{dt} \mathbf{w}^\intercal \delta\mathbf{x} \right)^{2} \right] \\
    &\propto  \Tr{\left( \mathbf{R}_\textrm{prep} \left( \left(\mathbf{L}_\textrm{prep} \, \mathbf{L}^{\intercal}_\textrm{prep} \right) \odot \boldsymbol{\Lambda} \right) \mathbf{R}^{\intercal}_\textrm{prep} \right)} + \beta \, N \; \mathbf{w}^\intercal \mathbf{R}_\textrm{prep} \left( \left(\mathbf{L}_\textrm{prep} \, \mathbf{L}^{\intercal}_\textrm{prep} \right) \odot \boldsymbol{\Gamma} \right) \mathbf{R}^{\intercal}_\textrm{prep} \nonumber
\end{align}
where $\mathbf{R}_\textrm{prep}$ and $\mathbf{L}_\textrm{prep}$ are the right and left eigenvectors of $\mathbf J_\textrm{prep}$, and its eigenvalues $\boldsymbol\lambda^{\textrm{prep}}$ are used to compute $\Lambda_{ij}=-1 / (\lambda^{\textrm{prep}}_i + \lambda^{\textrm{prep}}_j)$ and $\Gamma_{ij}=\lambda^{\textrm{prep}}_i\lambda^{\textrm{prep}}_j\Lambda_{ij}$. Finally, $N$ is the number of cortical units and $\beta$ is a hyperparameter which trades off the relative importance of transition speed and readout smoothness. Notice that this cost is not impacted by the shape of the initial rates' distribution.

\subsection{Simulation of the model confirms flexible, robust and extensible sequencing}

We simulated the model on the same task as the gradient-trained RNNs. The motif-specific parameters scale as in the multiplicative architecture (table~\ref{table_params}), but the thalamocortical model does have a few more hyperparameters. To make sure that these did not induce an inability to compare between approaches, we reduced the cortical size to $N=99$. After exploring a few values, we set $g=0.5\tau$, $K=10$, $P=50$,  $\beta=1/20$, and the motif transition duration to $5 \tau$. The readout weights $\mathbf{w}$ and recurrent weights $\mathbf{J^\mathrm{cc}}$ were sampled from a centered Gaussian distribution with std $1/\sqrt{N}$. The approximations $\hat{y}_\mu$ were fit to the target motifs $y_\mu$ under the constraints that: $\hat y_\mu(0)=0$; the elements of $\hat{\boldsymbol\lambda}_\mu$ had negative real part and were at least epsilon apart from each other; and the magnitudes of the elements of $\hat{\boldsymbol\alpha}_\mu$ were not exceedingly large. The resulting $\hat{\boldsymbol\lambda}_\mu$ and $\hat{\boldsymbol\alpha}_\mu$ were then used to optimize $\mathbf u_\mu$, $\mathbf v_\mu$, and $\mathbf b_\mu$ as described above.

We generated sequences from the thalamocortical model after initializing with iid standard Gaussian samples for the elements of $\mathbf x$ (this choice of a unit standard deviation indeed leads to readout values within the same range as the target motifs, Fig.~\ref{fig:thalamocort}c). Importantly, motifs were produced with low error either when starting from random initial conditions or when transitioning from another motif as part of a sequence (Fig.~\ref{fig:thalamocort}d,e). Of note, setting $K$ to 10 was a choice made to match the performance of the thalamocortical model to that of the gradient-trained ANNs in Fig.~\ref{fig:ANN_lim_robutsness} (see comparison in Fig.~\ref{fig:thalamocort}g). Larger $K$ would improve performance of this model.

\subsection{A thalamic-like transition module rescues transitioning in ANNs trained with SGD}\label{sec:prepwithSGD}

Due to the transition module, our thalamocortical model is asymptotically guaranteed to converge to the correct initial condition for the upcoming motif and tends to do so quickly if the eigenvalues are all large and negative (as in Fig.~\ref{fig:thalamocort}b). Hence, inspired by the success of this approach in guarding against transition failures, we revisited our additive and multiplicative networks and augmented their recurrent connectivity with the perturbation $\mathbf U_\textrm{prep}\mathbf V_{\textrm{prep}}^\intercal$ (with $P=50$), mimicking the preparatory dynamics of the thamalocortical model. We trained these models in two steps. First, with fixed $\mathbf J$ as described above and starting with standard normal iid entries for $\mathbf x$, but with no network input, we trained $\mathbf U_\textrm{prep}$ and $\mathbf V_{\textrm{prep}}$ with the cost function $\sum_t|\mathbf r(t)|^2$. This results in preparatory loop weights that drive the networks quickly to zero in $<10\tau$, similar to the linear case (see appendix and compare with Fig.~\ref{fig:thalamocort}b). Then we trained both networks on our 10 motifs (with no transitions) and modified training only as follows: for the first $5\tau$, we add our learned $\mathbf U_\textrm{prep}\mathbf V_{\textrm{prep}}^\intercal$ to the recurrent connectivity, and, emulating the thalamocortical model for the multiplicative network only, we use $\mathbf b_\mu$ and $\mathbf u_\mu\mathbf v_\mu^\intercal$ only during the preparatory and post-preparatory periods respectively. Both networks now learn better than before and, when instructed to generate sequences, show no transition failures (Figs.~\ref{fig:thalamocort}f,g).

\section{Discussion}

We trained RNNs with stochastic gradient descent on a flexible motif sequencing task and used random noise to leverage the generalization and interpolation capabilities of artificial networks~\cite{VezhnevetsFeudal2017,MordatchNIPS2015,MerelNeuralProbaMotorPrims2019}. These RNNs succeeded at learning randomly initialized individual motifs, but struggled to string motifs together in a sequence. This fundamentally boils down to a limitation of ANNs performance during out-of-distribution generalization, and is reminiscent of action transition issues in state-of-the-art networks during flexible movement control tasks~\cite{MerelHierarchVisuoMotor2019,Liu2012}. This issue could be circumvented by manually resetting the state of the network before each motif, however this would create undesirable discontinuities in the readout; alternatively, modules that are specific for each transition may be added, but this inefficient solution prevents 0-shot sequence improvisation, limiting flexibility. Using these tricks would become even more problematic in general settings where additional network modules process contextual inputs to flexibly select motifs with more levels of control hierarchy.

In contrast, these difficulties can be overcome by a network whose structure and dynamics are inspired by motor neuroscience: despite training motifs individually, this model can implement robust transitions while maintaining similar within-motif expressivity as the gradient-trained RNNs.

Our work outlines the need to constrain or design both the architecture and the dynamics of recurrent networks in order to achieve maximal performance, and calls for broadening the applications of the thalamocortical insights. For instance, the thalamocortical model could be extended with more modules and its dynamics could be smoothed by introducing and removing the connectivity perturbations in a gradual fashion, hence introducing more nonlinearities and functionalities while conserving some analytical tractability and performance guarantees. Alternatively, gradient-trained ANNs can be combined with a biologically-inspired motif transition module for robust sequencing, as we illustrate above and in more detail in the appendix. Therefore, our results open the door for improved recurrent network performance in real-world applications.

\section*{Broader Impact}

Our work lays theoretical foundations whose ultimate applications would involve flexible and extensible learning of complex continuous outputs. This is notably required to design robots that can be deployed in complex environments and learn from experience (for instance by observing another agent that performs a task~\cite{Argall2009}).

Ultimately, these robotics advances could be used in the manufacturing industry or domestically to free people from the necessity to perform actions themselves in order to obtain desired outcomes, such as producing goods or storing groceries. 

If this succeeds, many jobs performed by humans today could be performed by robots tomorrow. In order for such a change to benefit everyone rather than the robots' owners only, societal changes would need to occur such that the product of the robots' work would be fairly returned back to the whole society. Universal basic income may be a way to implement this redistribution. Moreover, psychosocial adjustments may be needed to adapt to the loss of some symbolic guidelines followed by many, such as the ideology promoting good character and hard work through promising monetary rewards in return and, more generally, the association of social status with income. 

\section*{Acknowledgements}
We thank Larry Abbott and Christopher J. Cueva for useful conversations. This research was supported by NIH BRAIN award (U19 NS104649), NSF/NIH Collaborate Research in Computational Neuroscience award (R01 NS105349), NIH Director’s Early Independence award (DP5 OD019897), and NSF NeuroNex award (DBI-1707398), as well as the Leon Levy foundation, the Gatsby Charitable Foundation, and the Swartz Foundation.

\bibliography{./main_submitted_Neurips}

\appendix
\pagebreak
\section*{Supplementary information}
\renewcommand{\thesubsection}{S.\arabic{subsection}}

\subsection{Motif generation}

For this study, we specifically choose motifs that challenge the dynamics of continuous time RNNs. Indeed, our motifs consist of a series of positive and negative discrete jumps with intervening constant periods. To generate a particular motif, we follow the following steps. First, we generate a centered Ornstein–Uhlenbeck process $z$ by running the dynamics $dz/dt = -z + \left( \gamma / \sqrt{dt} \right) \chi \left(t \right)$, where $\chi \left(t \right)$ is taken from a standard Gaussian independently at each timestep, dt=0.1 and $\gamma=3 \sqrt{2}$ such that after a pre-warming time the standard deviation of this process is 3. After a `warmup' period of $200 \, dt$ that we discard, we produce an additional $1000 \, dt$ of z-values that will be the basis for generating the motif. Second, we draw random time intervals $T_k$ with a uniform probability between $50 \, dt$ and $500 \, dt$ and select the first few in a list $[T_1, T_2, ..., T_j]$ such that $\sum_{k=1}^j T_k < 1000 \, dt$. Third, we set the value of the motif between the start and time $T_1$ to the average of the z-values over the same time interval; and similarly for the subsequent intervals the value of the motif between the times $[T_{i-1}+dt,T_i]_{i \in [2,...,k] }$ is set to the average of the z-values over the corresponding interval. Finally, we pad the end of the motif with zeros for $50dt$ and also reset the first value of the motif to 0.

\begin{figure}[h!]

\centering

\includegraphics[width=5.5in]{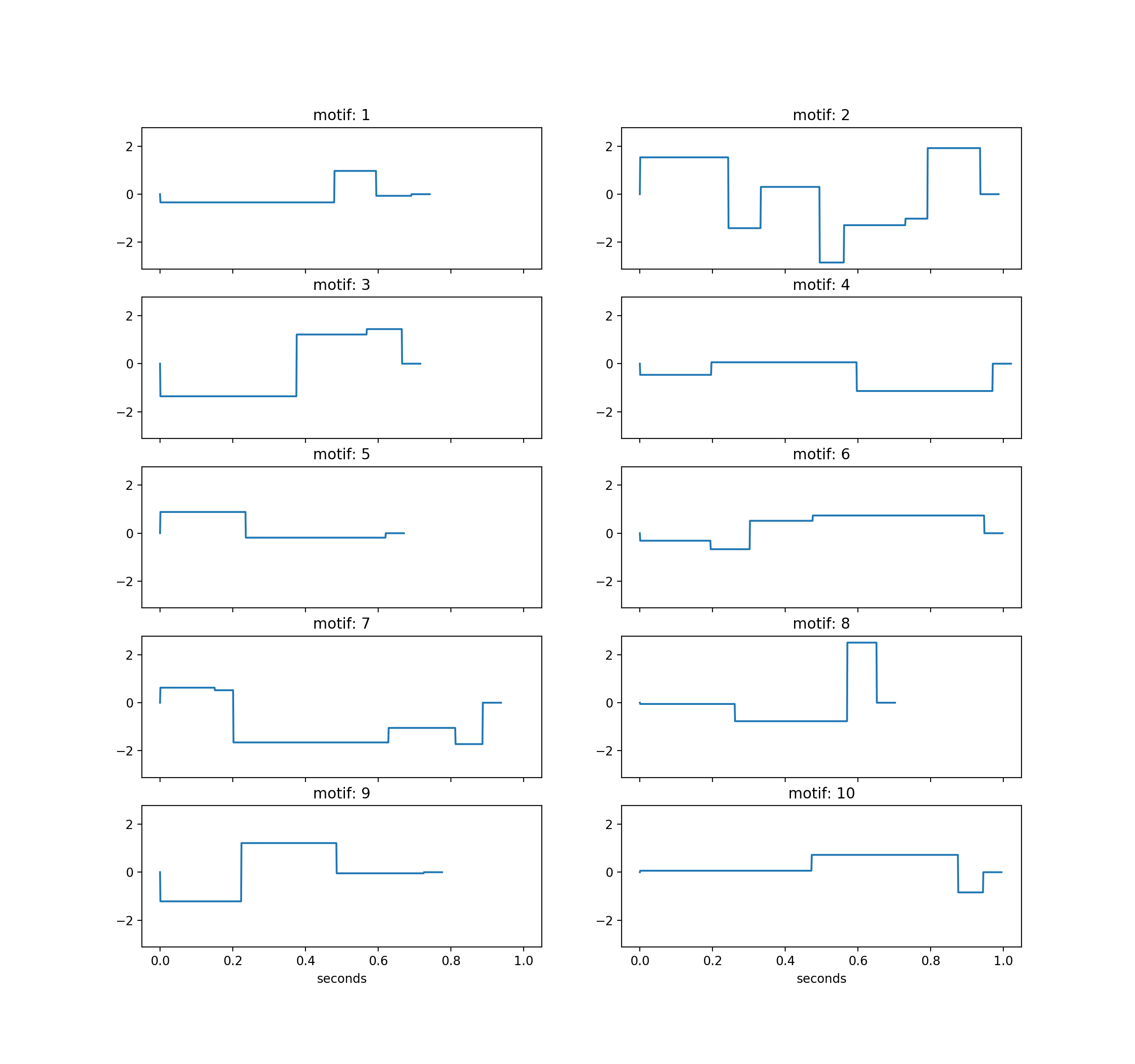}

\caption{\label{fig:motifs} \textbf{Motifs used in this paper}. All motifs were created according to the procedure discussed in the accompanying text.}
\end{figure}

Fig.~\ref{fig:motifs} lists the motifs used in this paper.

\subsection{Using a thalamic transition module with the additive and multiplicative networks \label{seq:ANNwithPrepaMeth}}

In Sec.~\ref{sec:prepwithSGD}, we presented the results of augmenting each of the additive and multiplicative networks with a 50-unit thalamic transition module that is active for the first $5\tau$ of every motif. Here we present a longer description of these results.

The dynamics of our nonlinear recurrent networks with the transition module but without any input are: $$\tau\dot{\mathbf x}=-\mathbf x+(g^x\mathbf J+\mathbf U_\textrm{prep}\mathbf V_\textrm{prep}^\intercal)\tanh(\mathbf x),$$ where $g^x$ is $g^\textrm{ad}$ or $g^\textrm{mu}$ for the additive and multiplicative networks respectively. The weights in $\mathbf U_\textrm{prep}$ and $\mathbf V_\textrm{prep}$ were initialized with centered Gaussian with std $\sqrt{0.05 / \sqrt{P * N}}$. We trained the weights of the transition modules with ADAM under the cost function $\sum_t|\mathbf r(t)|^2$. Minibatches consisted of 64 trials of length $20\tau$, each starting with random $\mathbf x$ values sampled iid from the standard normal distribution. After 1,000 minibatches, both $N=300$ and $N=100$ networks were seen to have converged.

\begin{figure}[t!]

\centering

\includegraphics[width=5.5in]{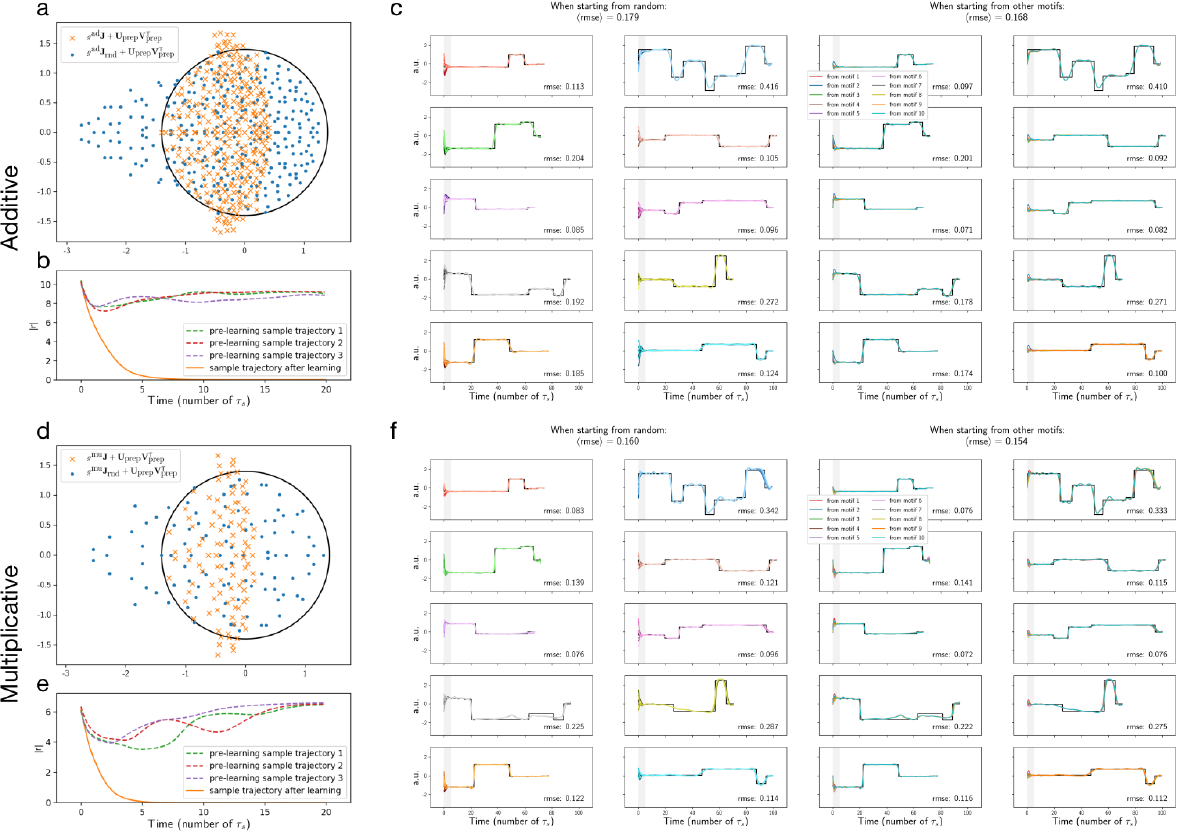}

\caption{\label{fig:nonlinear_with_prep}
\textbf{Using a thalamic transition module rescues transitioning for nonlinear RNNs trained with SGD}. \textbf{a,b,c} Additive model.
\textbf a. Eigenspectrum of $g^\textrm{ad}\mathbf J+\mathbf U_\textrm{prep}\mathbf V_\textrm{prep}^\intercal$ after training of $\mathbf U_\textrm{prep}$ and $\mathbf V_\textrm{prep}$ (orange crosses) and when replacing $\mathbf J$ with a random matrix not used during training (blue dots). Black circle has radius $g^\textrm{ad}$.
\textbf b. $|\mathbf r|$ versus time before (dotted lines) and after (solid line) training of $\mathbf U_\textrm{prep}$ and $\mathbf V_\textrm{prep}$.
\textbf c. Left: network outputs for
each motif when starting from 9 different random $\mathbf x$ values. Saturations (light to dark) indicate different trials. Right: network outputs for each motif when starting from the final $\mathbf x$ values of the other 9 motifs. Colors indicate the prior motif. The grey bars indicate the time during which the transition module was active.
\textbf{d,e,f}. As in \textbf{a,b,c} for the multiplicative model.
}
\end{figure}

Figs.~\ref{fig:nonlinear_with_prep}a,d show the eigenvalue distributions of $g^x\mathbf J+\mathbf U_\textrm{prep}\mathbf V_\textrm{prep}^\intercal$ after training. Importantly, though the thalamic module is low rank ($P<N$), all eigenvalues have real part significantly less than 1 which causes decay of the network rates towards a $\mathbf 0$ fixed-point in the vicinity of this fixed-point. For comparison, we show the eigenspectrum of $g^x\mathbf J_\textrm{rnd}+\mathbf U_\textrm{prep}\mathbf V_\textrm{prep}^\intercal$ for random matrix $\mathbf J_\textrm{rnd}$ which has the same statistics as $\mathbf J$. In this case, there continues to be large eigenvalues that will prevent fast rate decay. These results demonstrate that the solutions $\mathbf U_\textrm{prep}$ and $\mathbf V_\textrm{prep}$ needed to negate the amplifying dynamics of $\mathbf J$ are specific to that particular $\mathbf J$.

In Figs.~\ref{fig:nonlinear_with_prep}b,e, we show the time evolution of the norm of the rate vector $\mathbf r$ during three sample trajectories prior to learning and one post-learning (all post-learning samples are nearly identical). On the scale plotted, the norms after learning are indistinguishable from zero after approximately $7\tau$.

With the thalamic transition modules in hand, we retrained the additive and multiplicative networks as described in Secs.~\ref{seq:TasksModels} and \ref{sec:gradientRNNresults}. For the additive network, our only modification from before is that we included the thalamic transition module in the network dynamics for the first $5\tau$:
$$
\tau\dot{\mathbf x}=-\mathbf x+(g^\textrm{ad}\mathbf J+\mathbbm 1_{t\le5\tau}\mathbf U_\textrm{prep}\mathbf V_\textrm{prep}^\intercal)\tanh(\mathbf x)+\mathbf b_\mu.
$$
For the multiplicative network, to make a more direct comparison with the thalamocortical model presented in Sec.~\ref{sec:TCmodel}, we set the dynamics such that the input $\mathbf b_\mu$ and loop $\mathbf u_\mu\mathbf v_\mu^\intercal$ were only active during the transition and post-transition periods respectively:
$$
\tau\dot{\mathbf x}=-\mathbf x+(g^\textrm{mu}\mathbf J+\mathbbm 1_{t\le5\tau}\mathbf U_\textrm{prep}\mathbf V_\textrm{prep}^\intercal+\mathbbm 1_{t>5\tau}\mathbf u_\mu\mathbf v_\mu^\intercal)\tanh(\mathbf x)+\mathbbm 1_{t\le5\tau}\mathbf b_\mu.
$$
Figs.~\ref{fig:nonlinear_with_prep}c,f show that both networks learn to perform the task and show no degradation in their performance when tested on sequence generation (i.e., having initial $\mathbf x$ values given by the ends of other motifs) rather than when starting them with standard Gaussian $\mathbf x$ values (which was their training regime).

\subsection{Another strategy for incorporating insights from the preparatory period of the thalamocortical model into SGD-trained ANNs}

In the previous section, we described a successful approach for improving sequencing robustness that adds a module implementing all motif transitions to a $\tanh$ RNN. This succeeds because the design of this transition module is qualitatively different from the classical approaches using SGD to adjust the connection weights of RNNs. Indeed, this module is not trained on many possible sequence orders and randomness is not directly used to try to improve generalization across different transitions; instead, the transition module creates network dynamics that mimic the transition dynamics of the thalamocortical model. This approach has the advantage of intrinsically producing smooth dynamics and allowing more complex motif dynamics. However, asymptotic convergence guarantees are lost in this setting. Indeed, during a particular transition the network's dynamics (characterized by the connectivity $g^x\mathbf J+\mathbf U_\textrm{prep}\mathbf V_\textrm{prep}^\intercal$ and input $\mathbf{b}_{\mu}$ specific to the next motif) could have several fixed-points asides from the one which is reached in our simulations (Fig.~\ref{fig:nonlinear_with_prep}). Therefore, in a more general context where the activity of the network at the end of a motif would settle in an `exotic' part of the state-space far away from this fixed point, then the transition dynamics may not be able to bring the activities close to the appropriate pattern to start the next motif and the production of the next motif could be impaired.

Here, we would like to mention another training strategy that would retain asymptotic convergence guarantees during transitions while leveraging SGD training of a nonlinear network for motif production. Specifically, we suggest stitching a locally linear dynamical regime during transitions (i.e. now, during transitions only, $\mathbf x$ instead of $\tanh(\mathbf x)$ multiplies the synaptic weights in the dynamics' equations) with nonlinear dynamics during motif generation (i.e. the dynamics are then identical to those in section~\ref{seq:ANNwithPrepaMeth}). This permits setting the weights $\mathbf U_\textrm{prep}$ and $\mathbf V_\textrm{prep}$ using the same semi-analytical method as for the thalamocortical model, while training the motif-specific input and loop weights with SGD. There would then be a single stable fixed point for the dynamical regime during a given transition, whose value could be computed analytically as a function of the motif-specific input $\mathbf b_\mu$. This analytic solution for the transition dynamics' fixed point would allow us to modify the training of the motif-specific input $\mathbf{b}_{\mu}$ and/or loop weights $\mathbf u_\mu\mathbf v_\mu^\intercal$: in addition to training them on trials with a transition dynamics over a fixed duration (as in section~\ref{seq:ANNwithPrepaMeth}), there could be trials where a shortened `post-transition' motif should be produced with the initial activities $\mathbf x$ set to the transition dynamics' fixed point. Hence, this could guarantee the accuracy of the upcoming motif for a long-enough transition duration. Note that here, in contrast to the flexibility achieved in the fully-analytic thalamocortical model, increasing the transition duration would lead to a longer sequence duration as the beginning of a motif cannot be arbitrarily excised from the dynamics.
However, in a context when planning over several motifs is possible and there is a priori knowledge about how `hard' a given transition would be and therefore how much time should be spent transitioning, this limitation may be circumvented by designing a training protocol compatible with the placement of the transitioning dynamics during the end of the previous motif~\cite{Zimnik2020} instead of the beginning of the next motif as we did in this article.

\end{document}